\documentclass[sigconf]{acmart} 

\usepackage{tabularx}
\usepackage{multirow}
\usepackage{booktabs}
\usepackage{amsmath}
\usepackage{amsfonts}

\usepackage{mathtools}
\usepackage{amsthm}
\usepackage{array}
\usepackage{xcolor}
\usepackage[capitalize,noabbrev]{cleveref}

\AtBeginDocument{
  }

\renewcommand\footnotetextcopyrightpermission[1]{}
\setcopyright{none}
\acmDOI{}
\acmISBN{}
\acmConference[]{}{}{}
\acmYear{}
\copyrightyear{}

\begin{document}

\title{Spatial-Regularization-Aware Dual-Branch Collaborative Inference for Training-Free OVSS in Remote Sensing Imagery}

\author{Jianzheng Wang}
\affiliation{%
  \institution{Nanjing University of Information Science \& Technology}
  \city{Nanjing}
  \country{China}
}

\author{Huan Ni}
\authornote{Corresponding Author.}
\affiliation{%
  \institution{Nanjing University of Information Science \& Technology}
  \city{Nanjing}
  \country{China}
}
\email{nih@nuist.edu.cn}

\renewcommand{\shortauthors}{Wang and Ni}

\begin{abstract}
High-resolution remote sensing images contain densely distributed objects with pronounced scale variations and complex boundaries, which impose higher demands on both the geometric localization and semantic prediction capabilities of semantic segmentation models. Existing training-free open-vocabulary semantic segmentation (OVSS) methods typically fuse Contrastive Language-Image Pretraining (CLIP) and vision foundation models (VFMs) using “one-way injection” and “shallow post-processing” strategies, making it difficult to satisfy these requirements. To address this issue, we propose a spatial-regularization-aware dual-branch collaborative inference framework for training-free OVSS, termed SDCI. First, during feature encoding, SDCI introduces a cross-model attention fusion (CAF) module, which guides collaborative inference by injecting self-attention maps into each other. Second, we propose a bidirectional cross-graph diffusion refinement (BCDR) module that enhances the reliability of dual-branch segmentation scores through iterative random-walk diffusion. Finally, we incorporate low-level superpixel structures and develop a convex-optimization-based superpixel collaborative prediction (CSCP) mechanism to further refine object boundaries. Experiments on multiple remote sensing semantic segmentation benchmarks demonstrate that our method achieves better performance than existing approaches. Our code is available at https://github.com/yu-ni1989/SDCI. 
\end{abstract}



\maketitle

\section{Introduction}
\label{SEC_INTRODUCTION}
Vision-language models (VLMs) and vision foundation models (VFMs) have been leveraged to meet the objective of open-vocabulary semantic segmentation (OVSS) ~\cite{Ge2025CRTNet}. Existing approaches can be classified as training-based~\cite{Zhong2025OmniSAM} and training-free~\cite{Shi2025Trident, Zhang2025E-SAM} methods. Compared with training-based approaches, training-free methods are more flexible, and require no expensive manual annotations, thus showing great application potential. However, extending training-free OVSS techniques to the remote sensing domain poses unique challenges. Compared with natural images, high-resolution remote sensing imagery typically exhibits a top-down ``bird's-eye view'' perspective, characterized by densely distributed ground objects, drastic scale variations, and extremely complex boundaries. Such differences imply that remote sensing tasks impose stricter requirements on both geometric localization and semantic descriptions. 

On the one hand, as a representative VLM, the Contrastive Language-Image Pretraining (CLIP) \cite{Radford2021CLIP} which can generate high-level semantics in a zero-shot setting by leveraging text prompts, is widely used. To meet the objective of OVSS, existing works mainly focus on modifying CLIP's internal attention mechanism, such as ClearCLIP~\cite{Lan2025ClearCLIP} and CLIPer~\cite{Sun2025CLIPer}. However, CLIP is mainly designed for image-level classification, and thus lacks pixel-level spatial localization ability, leading to the difficulty for precise geometric positional information (such as pixel-level boundaries) acquirement in remote sensing tasks. On the other hand, VFMs such as DINO \cite{Caron2021DINO} and SAM~\cite{Kirillov2023SAM} have been widely adopted in foreground object segmentation tasks~\cite{Simeoni2021,Simeoni2022,Wang2022}. Although VFMs excel at modeling spatial details, their learned embedding spaces are not explicitly aligned with textual semantics, and thus they lack semantic description capability tailored to remote sensing tasks. To capitalize on the strengths of CLIP and VFMs while mitigating their shortcomings, some studies have begun to explore integrating them. For example, ProxyCLIP~\cite{Lan2024ProxyCLIP} borrows attention maps from external VFMs to guide CLIP, and CorrCLIP~\cite{Zhang2025CorrCLIP} leverages VFMs to estimate the interaction range of similarity-based patch relations, thereby reducing inter-class correlation. We summarize this line of work as a ``one-way injection" from VFMs to CLIP, or from CLIP to VFMs. In addition, some approaches conduct fusion in the output space using masking~\cite{Barsellotti2024bFreeDA}, graph matching~\cite{Kim2025CASS}, or the strategy that employing independently pretrained modules~\cite{Li2025SegEarth}. These methods typically treat VFMs or vision operators as a static ``black box", using only their outputs for fusion and post-processing. We summarize this line of work as ``shallow post-processing". The ``one-way injection" and ``shallow post-processing" fail to achieve deep, bidirectional fusion, leading to inferior results in segmenting remote sensing images.

\begin{figure}[ht]
\begin{center}
\centerline{\includegraphics[width=3.2in]{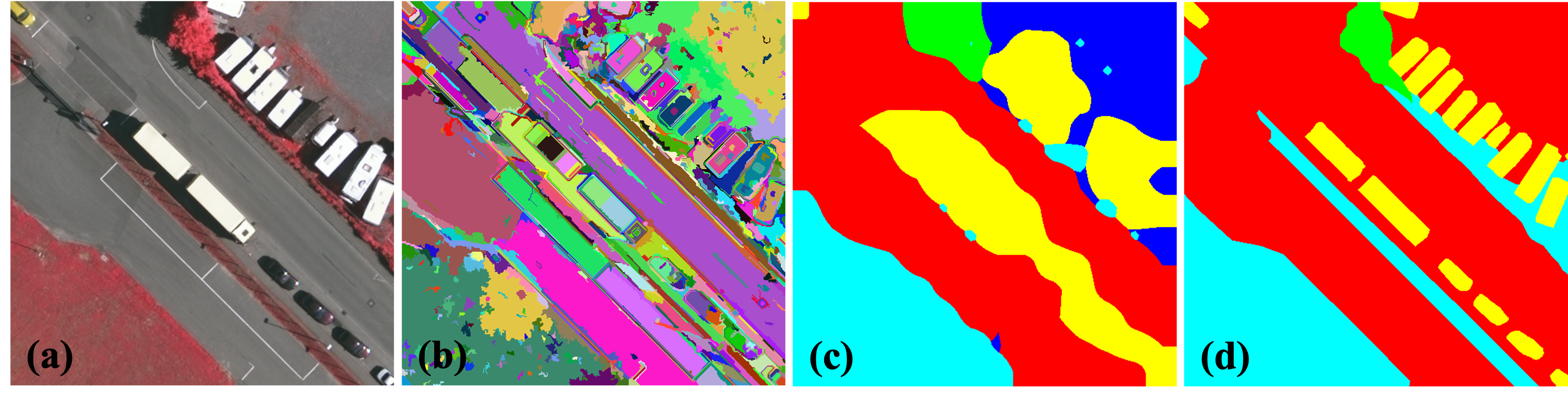}}
\caption{Observation on the boundary-enhancing effect of superpixels. (a) original remote sensing image, (b) superpixel segmentation result; (c) semantic segmentation result generated by CLIPer, (d) ground truth.}
\label{FIG_MOTIVATION2}
\end{center}
\end{figure}

Meanwhile, we observe that superpixel structures naturally enhance object boundaries, which can effectively compensate for CLIP’s limitations in pixel-level spatial localization ability in remote sensing tasks. As illustrated in Fig.~\ref{FIG_MOTIVATION2}(a)--(b), the superpixels closely adhere to the physical contours of objects; in contrast, the segmentation produced by CLIPer~\cite{Sun2025CLIPer} in Fig.~\ref{FIG_MOTIVATION2}(c) remains blurry near boundaries and fails to precisely distinguish different land-cover categories. This comparison indicates that the low-level geometric priors carried by superpixels are crucial for addressing the ``blurry-boundary'' issue. 


On top of these observations, we propose a new spatial regu\-lar\-ization-aware dual-branch collaborative inference framework for OVSS in a training-free manner, termed SDCI. SDCI consists of three core collaborative components, i.e., a cross-model attention fusion (CAF) module and a bidirectional cross-graph diffusion-based refinement (BCDR) module, together with a convex-optimization-based superpixel collaborative prediction (CSCP) mechanism. Specifically, CAF performs bidirectional injection of attention maps during feature encoding, enabling CLIP's semantic predictions to be propagated with DINO's structural information, while simultaneously enriching DINO's structural perception with CLIP's semantic cues. Based on this interaction, we obtain high-quality initial segmentation results. Then, BCDR places the initial results into a global refinement process via iterative random-walk diffusion: it uses the ``structural graph'' constructed by DINO to correct spatial inconsistencies in CLIP, and uses the ``semantic graph'' constructed by CLIP to merge semantically separated regions in DINO. Finally, considering the complex boundaries of ground objects and the high geometric-precision requirement in remote sensing imagery, we introduce superpixel structures and convex optimization to deeply fuse the initial segmentation outputs from both the CLIP and DINO branches. Extensive experiments demonstrate the effectiveness of our method, and our contributions are summarized as follows:
\begin{itemize}
\item CAF and BCDR that enable bidirectional interaction between CLIP’s semantic knowledge and DINO’s structural information are proposed. The two modules address both the ``one-way injection" and ``shallow post-processing" issues, exchanging complementary information and mitigating noise interference. 
\item CSCP that formulates a global energy minimization framework and introduces superpixels as a strong geometric constraint is proposed. CSCP deeply integrates the semantic probability predictions produced by CLIP and DINO with low-level superpixel topology to precisely sharpen segmentation boundaries, which further addresses the ``shallow post-processing" issue.  
\item Building on the above modules, we present a novel training-free collaborative inference framework, SDCI. By deeply fusing information at both the feature encoding stage and the segmentation refinement stage, the proposed framework hierarchically improves OVSS performance, achieving superior results over existing methods on multiple remote sensing semantic segmentation benchmarks.
\end{itemize}

\section{Related Work}
\label{SEC_RELATEDWORK}
\subsection{Methods based on improving internal mechanisms of CLIP}
The central idea of this line of research is to exploit and rectify the attention mechanisms within CLIP itself, making it suitable for pixel-level tasks~\cite{Hajimiri2024NACLIP, Shao2025CLIPtrase, Sun2023CaR, Rao2021DenseCLIPLD}. For example, MaskCLIP~\cite{Zhou2022MaskCLIP} employs an identical self-self matrix as the self-attention map at last layer to generate visual patch embeddings. SCLIP~\cite{Wang2025SCLIP} and ClearCLIP~\cite{Lan2025ClearCLIP} introduce self--self attention variants to replace the original query--key attention, thereby improving the spatial consistency of feature maps. Different from those approaches that completely discard query--key interactions, ResCLIP~\cite{Yang2024ResCLIP} extracts cross-correlation features preserved in intermediate layers and restores the spatial localization capability of the last layer via residual connections. However, as pointed out by the fine-grained analysis in clip-oscope~\cite{Abbasi2025ClipOscope}, CLIP exhibits significant biases toward large objects and specific text ordering in multi-object scenes. Such inherent deficiencies in spatial and structural perception make it difficult for purely internal correction methods in CLIP to achieve accurate boundary localization in remote sensing scenarios.

\subsection{Methods using only VFMs}
Leveraging the robust visual representations of DINO series~\cite{Caron2021DINO,Oquab2024DINO-v2,Simeoni2025DINO-v3}, researchers have developed several unsupervised segmentation methods. STEGO~\cite{hamilton2022unsupervised} distills dense feature correspondences into high-quality discrete semantic labels. TokenCut~\cite{Wang2023TokenCut} constructs a fully connected graph based on image features and segments salient objects using the Normalized Cut algorithm~\cite{Shi2000NCuts}. CutLER~\cite{Wang2023CutLER, Wang2024VCutLER} extends this idea by discovering multiple objects from self-supervised features, enabling zero-shot unsupervised localization. Meanwhile, SAM series~\cite{Kirillov2023SAM, Ravi2024sam2} demonstrates remarkable category-agnostic segmentation capability. Subsequently, improved SAM-based methods, such as E-SAM~\cite{Zhang2025E-SAM} and OmniSAM~\cite{Zhong2025OmniSAM}, have enhanced segmentation completeness and accuracy to varying degrees. However, the ``structure without semantics'' property prevents them from independently solving OVSS, highlighting the necessity of collaborating with CLIP.

\subsection{Multi-model collaborative methods}
Firstly, some approaches attempt to support segmentation by constructing external data assistance, such as ReCo~\cite{Shin2022ReCo}, OVDiff~\cite{Karazija2025OVDiff}, FOSSIL~\cite{Barsellotti2024aFOSSIL}, and FreeDA~\cite{Barsellotti2024bFreeDA}. However, these methods need large-scale retrieval and offline generation~\cite{Rombach2022LDM}, which introduces substantial computational overhead and pipeline complexity. Other approaches focus on aligning features across models by employing learnable parameters~\cite{Barsellotti2025Talk2DINO, Wysoczanska2025CLIPDINOiser, Wang2025}.Recent multi-model collaborations also tackle complex instruction-following tasks; for instance, CORA~\cite{Howlader_2026_WACV} achieves reasoning segmentation by explicitly fine-tuning a CLIP-LLM-SAM pipeline. Although effective, introducing learnable parameters sacrifices the pure training-free flexibility.

In contrast, some works explore fully training-free strategies using both CLIP and VFMs~\cite{Kang2024LaVG, Lan2024ProxyCLIP, Zhang2025CorrCLIP, zhang2023personalize}. For instance, PI-CLIP~\cite{Wang2024PICLIP} generates high-quality prior information for segmentation by leveraging the zero-shot visual-text alignment capacity of CLIP. Similarly, VLG\_SAM~\cite{Yoon_2026_WACV} guides SAM via vision-language prompts for training-free segmentation. CASS~\cite{Kim2025CASS} further proposes injecting features from DINO into the attention mechanism of CLIP to enhance object-level contextual consistency. SegEarth-OV~\cite{Li2025SegEarth} improves FeatUp~\cite{Fu2024SimFeatUp} and introduces the independently pretrained SimFeatUp for post-processing. However, these methods often rely on ``one-way information injection" or ``shallow post-processing", and fail to achieve deep fusion between the two models.

\section{Methodology}
\label{SEC_METHODOLOGY}

\begin{figure*}[ht]

\begin{center}
\centerline{\includegraphics[width=6in]{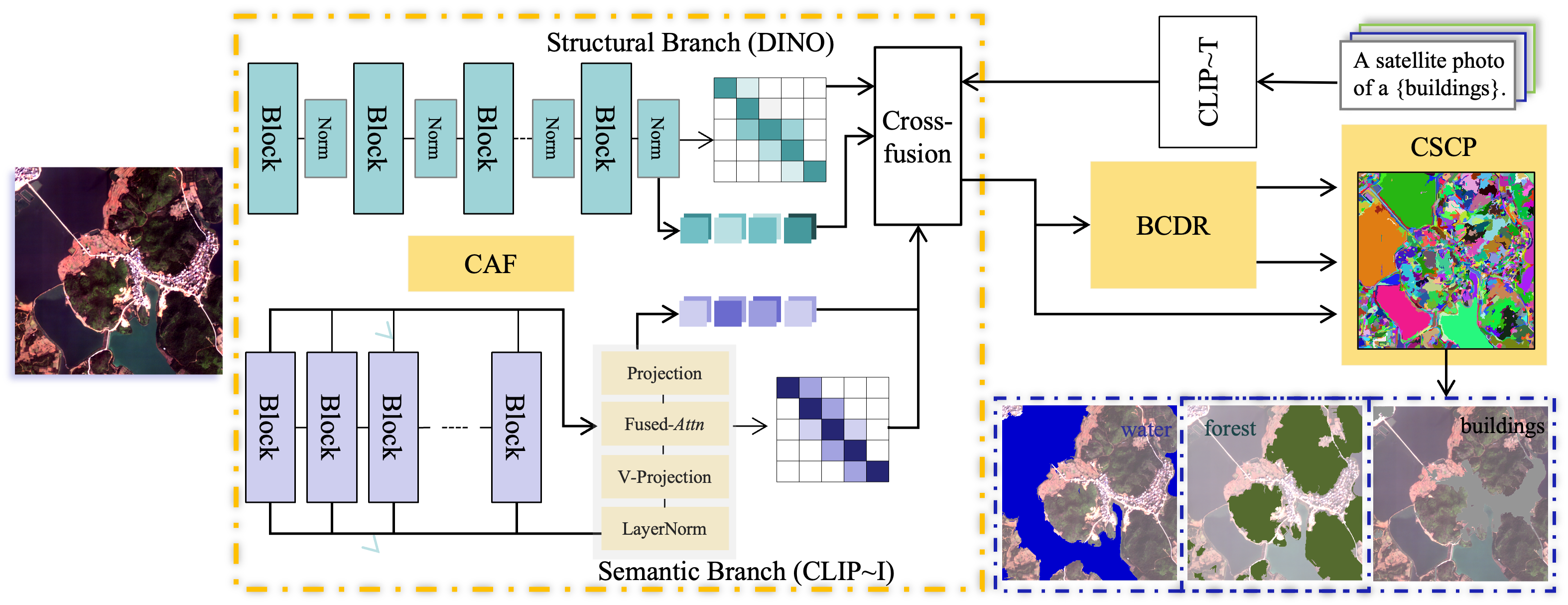}}
\caption{The proposed SDCI framework, in which CAF, BCDR, and CSCP serve as the core modules. CAF enables interaction between semantic and structural information by exchanging attention maps. The resulting initial logit maps are then fed into the BCDR module, which performs cross diffusion between the structural graph constructed by DINO and the semantic graph constructed by CLIP to achieve globally consistent enhancement. Finally, CSCP imposes constraints from superpixel structures and, through a convex optimization process, fuses the predictions from the two branches to generate the final segmentation map.}
\label{FIG_OVERVIEW}
\end{center}

\end{figure*}

\subsection{Overview}
Figure \ref{FIG_OVERVIEW} illustrates the overall architecture of SDCI. SDCI mainly consists of three core components: CAF, BCDR and CSCP.  

\subsection{CAF}
\label{SEC_CAF}
CAF first processes the input image in parallel using the CLIP image encoder and DINO, and then performs cross-fusion on top of them.

\paragraph{Semantic Branch.}
To enhance the spatial consistency of patch embeddings, we adopt a multi-level feature fusion strategy similar to CLIPer~\cite{Sun2025CLIPer}. Given the patch embedding $F_{0,\mathrm{clip}} \in \mathbb{R}^{(hw+1)\times D}$, we feed it into a stack of Transformer blocks. For the $n$-th Transformer block, the output embedding $F_{n,\mathrm{clip}}$ is computed via standard self-attention $\mathrm{Att}(\cdot)$ and feed-forward networks $\mathrm{FFN}(\cdot)$:
\begin{equation}
F'_{n,\mathrm{clip}}=\mathrm{Att}(Q_n, K_n,V_n)+F_{n-1,\mathrm{clip}},
\label{eq:clip_1}
\end{equation}
\begin{equation}
F_{n,\mathrm{clip}}=\mathrm{FFN}\!\left(\mathrm{LN}\!\left(F'_{n,\mathrm{clip}}\right)\right)+F'_{n,\mathrm{clip}},
\label{eq:clip_2}
\end{equation}
where $Q_n$, $K_n$, and $V_n$ denote the query, key, and value matrices obtained by linearly projecting $F_{n-1,\mathrm{clip}}$. $\mathrm{LN}(\cdot)$ is a normalization layer. The attention map at each layer is computed as
\begin{equation}
A_n=\mathrm{Softmax}\!\left(\frac{Q_n K_n^{\top}}{\sqrt{d_k}}\right),
\label{eq:clip_3}
\end{equation}
where $d_k$ is the feature dimension of each attention head. To leverage the rich spatial information from early layers, we collect and average the attention maps of all intermediate layers to obtain an averaged attention map:
\begin{equation}
A_{\mathrm{avg}}=\frac{1}{N-1}\sum_{n=1}^{N-1} A_n.
\label{eq:avg_4}
\end{equation}
We then use $A_{\mathrm{avg}}$ as a spatial prior to guide the feature processing in the $N$-th layer:
\begin{equation}
F_{N,\mathrm{clip}}=\mathrm{Norm}\!\Big(\mathrm{ReLU}\!\big(\mathrm{Sym}(A_{\mathrm{avg}})-\mu\big)\Big)\cdot V_N,
\label{eq:clip_5}
\end{equation}
where $\mathrm{Sym}(\cdot)$ denotes the symmetrization operation, and $\mu$ is the mean value of $A_{\mathrm{avg}}$. $\mathrm{Norm}(\cdot)$ denotes $\ell_{1}$ normalization, and $\mathrm{ReLU}(\cdot)$ sets negative values to zero. This process yields the final feature embeddings $\{F_{1,\mathrm{clip}}, \ldots, F_{N,\mathrm{clip}}\}$ for the semantic branch. Moreover, we average $A_{\mathrm{avg}}$ across the head dimension to obtain the effective attention map $A_{N,\mathrm{clip}}$:
\begin{equation}
A_{N,\mathrm{clip}}=\frac{1}{H}\sum_{h=1}^{H} A_{\mathrm{avg}}^{(h)},
\label{eq:clip_6}
\end{equation}
where $H$ is the number of heads in multi-head attention, and $A_{\mathrm{avg}}^{(h)}$ denotes the component of $A_{\mathrm{avg}}$ in the $h$-th head.

\paragraph{Structural Branch.}
In parallel, we process the same input image using the DINO visual encoder. Different from the fusion strategy adopted in the semantic branch, our strategy for the structural branch is to maximally preserve its original high-quality hierarchical structural information. For the output feature $F'_{n,\mathrm{dino}}$ from the $n$-th Transformer block, before collecting it, we pass it through a normalization layer to obtain the normalized feature $F_{n,\mathrm{dino}}$, which can be formulated as
\begin{equation}
F_{n,\mathrm{dino}}=\mathrm{Norm}\big(F'_{n,\mathrm{dino}}\big),
\quad \forall n\in\{1,\cdots,N\}.
\label{eq:caf_dino_1}
\end{equation}
This ensures that features from low to high layers are mapped into a normalized and stable feature space. We then stack these normalized hierarchical features to form a full-level feature tensor $\{F_{1,\mathrm{dino}},\cdots,F_{N,\mathrm{dino}}\}$. Meanwhile, we extract the standard self-attention map from the last Transformer block, denoted as $A_{N,\mathrm{dino}}$.

\paragraph{Cross-Fusion.}
After extracting the hierarchical features and attention maps from both branches via the above asymmetric strategies, knowledges are fused. First, we spatially align the final effective attention maps of the two branches. Meanwhile, we compare the full-level features $\{F_{1,\mathrm{clip}},\cdots,F_{N,\mathrm{clip}}\}$ and $\{F_{1,\mathrm{dino}},\cdots,F_{N,\mathrm{dino}}\}$ with the text embedding $E_T$, obtaining the corresponding full-level segmentation logits $\{S_{1,\mathrm{clip}},\cdots,S_{N,\mathrm{clip}}\}$ and $\{S_{1,\mathrm{dino}},\cdots,S_{N,\mathrm{dino}}\}$. Notably, although the feature space of DINO is not explicitly aligned with text, the intrinsic structure emerging from its high-quality visual representations enables the comparison with $E_T$ to still produce meaningful initial segmentation results rich in spatial details.

Next, we compute the final preliminary segmentation scores using a cross-fusion formulation. For the semantic branch, the enhanced score $S'_{\mathrm{clip}}$ is computed as
\begin{equation}
\begin{split}
S'_{\mathrm{clip}}
=
\Big(\mathrm{Align}(A_{N,\mathrm{clip}})
+\lambda_1\cdot \mathrm{Align}(A_{N,\mathrm{dino}})\Big)\\\cdot S_{N,\mathrm{clip}}
+\frac{1}{N-1}\sum_{n=1}^{N-1}S_{n,\mathrm{clip}}.
\end{split}
\label{eq:caf_fuse_1}
\end{equation}
Here, $\mathrm{Align}(\cdot)$ denotes a spatial alignment operator, and $\lambda_1$ is a hyperparameter used to balance the strength of structured guidance from DINO (we set $\lambda_1=1$ by default). The terms $S_{N,\mathrm{clip}}$, $S_{n,\mathrm{clip}}$ and $A_{N,\mathrm{clip}}$ are obtained from the semantic branch described above.

Similarly, the cross-fusion for the structural branch follows the similar way, and the produced score $S'_{\mathrm{dino}}$ is computed as
\begin{equation}
\begin{split}
S'_{\mathrm{dino}}
=
\Big(\mathrm{Align}(A_{N,\mathrm{dino}})
+\lambda_1\cdot \mathrm{Align}(A_{N,\mathrm{clip}})\Big)\\\cdot S_{N,\mathrm{dino}}
+\frac{1}{N-1}\sum_{n=1}^{N-1}S_{n,\mathrm{dino}}.
\end{split}
\label{eq:caf_fuse_2}
\end{equation}
The inputs $S_{N,\mathrm{dino}}$, $S_{n,\mathrm{dino}}$ and $A_{N,\mathrm{dino}}$ are obtained from the structural branch, while all other symbols and parameters share the same definitions as those in Eq.~\ref{eq:caf_fuse_1}. Through CAF that combines high-level guidance with full-level aggregation, our method achieves a favorable balance between the accuracy of semantic decision-making and the fidelity of spatial details, producing high-quality initial segmentation results for the subsequent BCDR.

\subsection{BCDR}
\label{SEC_BCDR}
To achieve global enhancement of the preliminary segmentation logits, BCDR first explicitly models the discrete image patches in the spatial domain as a graph structure. Furthermore, since different models perceive different intrinsic relationships within an image, we can construct a semantic graph and a structural graph, which respectively serve different objectives for the enhancement.

\paragraph{Semantic Graph Construction.}
The semantic graph is constructed from the semantic branch. Given the CLIP features $\{F_{1,\mathrm{clip}},\cdots,F_{N,\mathrm{clip}}\}$, we compute
\begin{equation}
F_{\mathrm{clip}}=\frac{1}{N}\sum_{i=1}^{N}F_{i,\mathrm{clip}}.
\label{eq:bcdg_sem_1}
\end{equation}
After applying $\ell_2$ normalization, each vector in $F_{\mathrm{clip}}$ has unit norm, which ensures that $F_{\mathrm{clip}}F_{\mathrm{clip}}^{\top}$ becomes a pairwise cosine-similarity matrix. Accordingly, the transition matrix $T_{\mathrm{clip}}$ is defined as
\begin{equation}
T_{\mathrm{clip}}
=
\mathrm{Norm}_{\mathrm{row}}
\!\left(
\exp\!\left(
\frac{S_K\!\left(F_{\mathrm{clip}}F_{\mathrm{clip}}^{\top}\right)}{\tau}
\right)
\right).
\label{eq:bcdg_sem_2}
\end{equation}
Here, the cosine similarities in $F_{\mathrm{clip}}F_{\mathrm{clip}}^{\top}$ define the affinities between nodes. The operator $S_K(\cdot)$ denotes a $K$-nearest-neighbor sparsification, which retains only the top-$K$ non-diagonal similarities in each row and sets all remaining entries to $-\infty$. The parameter $\tau$ is a temperature scaling coefficient that controls the sharpness of affinity weights, and is set to $0.07$ by default in this work. The function $\exp(\cdot)$ converts the sparsified similarities into non-negative affinity weights. Finally, $\mathrm{Norm}_{\mathrm{row}}(\cdot)$ denotes row-wise normalization, ensuring that each row sums to $1$.

\paragraph{Structural Graph Construction.}
Based on the DINO features $\{F_{1,\mathrm{dino}},\allowbreak \cdots,F_{N,\mathrm{dino}}\}$, we define the structural graph transition matrix $T_{\mathrm{dino}}$, whose edge weights strongly reflect pixel-level physical connectivity, as
\begin{equation}
T_{\mathrm{dino}}
=
\mathrm{Norm}_{\mathrm{row}}
\!\left(
\exp\!\left(
\frac{S_K\!\left(F_{\mathrm{dino}}F_{\mathrm{dino}}^{\top}\right)}{\tau}
\right)
\right).
\label{eq:bcdg_str_1}
\end{equation}
The definitions of all symbols follow the same procedure as in the computation of $T_{\mathrm{clip}}$.

\paragraph{Bidirectional refinement.}
Given $T_{\mathrm{clip}}$ and $T_{\mathrm{dino}}$, BCDR designs a novel cross-graph diffusion mechanism to perform global refinement with complementary strengths. This mechanism smooths the preliminary segmentation logits via an iterative random-walk process. Specifically, the refinement is bidirectional and symmetric: we use $T_{\mathrm{dino}}$ to refine the CLIP semantic scores $S'_{\mathrm{clip}}$, and use $T_{\mathrm{clip}}$ to refine the DINO structural scores $S'_{\mathrm{dino}}$.

For the semantic branch, the scores after $t$ diffusion steps, denoted as $S_{\mathrm{clip}}^{(t)}$, are computed as
\begin{equation}
S_{\mathrm{clip}}^{(t)}
=
\alpha \, T_{\mathrm{dino}} \, S_{\mathrm{clip}}^{(t-1)}
+
(1-\alpha)\, S'_{\mathrm{clip}}.
\label{eq:bcdg_br_1}
\end{equation}
Symmetrically, for the structural branch, the scores after $t$ diffusion steps, denoted as $S_{\mathrm{dino}}^{(t)}$, are computed as
\begin{equation}
S_{\mathrm{dino}}^{(t)}
=
\alpha \, T_{\mathrm{clip}} \, S_{\mathrm{dino}}^{(t-1)}
+
(1-\alpha)\, S'_{\mathrm{dino}}.
\label{eq:bcdg_br_2}
\end{equation}
Here, $\alpha\in(0,1)$ is a smoothing factor that balances neighborhood propagation and the initial information, and we set $\alpha=0.9$ by default in this work. We set the total diffusion steps $T$ to $10$, and the final enhanced scores are defined as
\begin{equation}
S_{g\text{-}\mathrm{clip}} = S_{\mathrm{clip}}^{(T)}, 
\qquad
S_{g\text{-}\mathrm{dino}} = S_{\mathrm{dino}}^{(T)}.
\label{eq:bcdg_br_3}
\end{equation}

In this way, we enforce the CLIP semantic predictions to propagate along the accurate physical paths depicted by DINO, thereby enhancing spatial consistency. Meanwhile, we leverage the CLIP semantic graph to merge regions in the DINO predictions that may be separated due to appearance variations, ensuring semantic completeness. Equations~\eqref{eq:bcdg_br_1}--\eqref{eq:bcdg_br_3} finally output two enhanced scores, namely $S_{g\text{-}\mathrm{clip}}$ and $S_{g\text{-}\mathrm{dino}}$, providing higher-quality inputs for the final CSCP stage.

\subsection{CSCP}
\label{SEC_CSCP}
To enforce the final segmentation map to be locally consistent with the low-level natural structures of the image, while also fusing the predictions from the two parallel branches, we propose a convex optimization-based superpixel collaborative prediction mechanism, namely CSCP.

First, we adopt the Felzenszwalb algorithm~\cite{Felzenszwalb2004} to decompose the input image $I$ into a set of superpixels $S=\{s_1,\cdots,s_M\}$. Then, using the topology of these superpixels, we construct a graph structure $W$ with adaptive edge weights. Specifically, for any adjacent pixel pair $(p,q)$ in the image, the connection weight $w_{p,q}$ is defined as
\begin{equation}
w_{p,q}
=
w_{\mathrm{cross}}
+
\big(w_{\mathrm{in}}-w_{\mathrm{cross}}\big)\cdot \mathbb{I}\big(s(p)=s(q)\big),
\label{eq:cscp_1}
\end{equation}
where $s(\cdot)$ denotes the superpixel index mapping such that, for a pixel $p$, if $p\in s_i$ (i.e., pixel $p$ belongs to the $i$-th superpixel in $S$), then $s(p)=i$. The function $\mathbb{I}(\cdot)$ is an indicator function that equals $1$ when the condition holds and $0$ otherwise. In our experiments, we set $w_{\mathrm{in}}=1.0$ and $w_{\mathrm{cross}}=0.10$. Within each superpixel, pixels are assigned larger connection weights to impose smoothness constraints; across superpixel boundaries, the weights are significantly reduced, allowing gradient discontinuities in the segmentation map and thus preserving object contours.

Next, we take the globally refined predictions from BCDR, namely the CLIP output $S_{g\text{-}\mathrm{clip}}$ and the DINO output $S_{g\text{-}\mathrm{dino}}$, as inputs. We construct a global energy function $E(Q)$, aiming to find an optimal probability distribution $Q$ that satisfies both semantic consistency and geometric smoothness. The optimization problem is formulated as
\begin{equation}
\begin{split}
\min_{Q\in\Delta} E(Q)
= 
\sum_{p\in\Omega}
[\lambda_C\,\mathrm{KL}\!\left(Q_p\,\|\,S_{g\text{-}\mathrm{clip}}(p)\right)+\\
\lambda_D\,\mathrm{KL}\!\left(Q_p\,\|\,S_{g\text{-}\mathrm{dino}}(p)\right)+\\
\beta
\sum_{(p,q)\in\varepsilon}
w_{p,q}\,\|Q_p-Q_q\|_{1}],
\end{split}
\label{eq:cscp_2}
\end{equation}
where $Q$ is the target refined probability distribution to be optimized, and $Q_p\in\mathbb{R}^{K}$ denotes the $K$-dimensional class probability vector at pixel $p$. The set $\Omega$ represents the image domain (i.e., the set of all pixels). The symbol $\Delta$ denotes the simplex constraint, defined as
\begin{equation}
\Delta
=
\left\{
Q \,\bigg|\,
\sum_{k=1}^{K} Q_p^{k}=1,\;
Q_p^{k}\ge 0
\right\},
\label{eq:cscp_3}
\end{equation}
which guarantees that the optimized result satisfies the mathematical properties of a valid probability distribution. $\mathrm{KL}(\cdot\|\cdot)$ denotes the Kullback--Leibler divergence, which measures the information discrepancy between the refined distribution $Q$ and the original predictions. The coefficients $\lambda_C$ and $\lambda_D$ are balancing factors that adjust the confidence assigned to the two branch predictions; in our experiments, we set $\lambda_C=1.0$ and $\lambda_D=0.2$. The parameter $\beta$ is a regularization weight that controls the overall strength of spatial smoothness. The set $\varepsilon$ contains all adjacent pixel pairs on the image grid. The adaptive weight $w_{p,q}$ is generated from the superpixel prior using Eq.~\ref{eq:cscp_1}. The norm $\|\cdot\|_{1}$ denotes the $\ell_1$ norm. Due to that the above energy function $E(Q)$ contains a non-smooth total variation term, we adopt the Primal--Dual Hybrid Gradient (PDHG) algorithm~\cite{Chambolle2011PDHG} to solve it iteratively, which guarantees convergence to the global optimum.

\section{Experiment}
\label{SEC_EXPERIMENT}
To comprehensively evaluate the performance of SDCI across different sensors, spatial resolutions, and top-view scenarios, we conduct experiments on a range of widely used yet highly challenging remote sensing semantic segmentation datasets, including GID~\cite{Tong2020GID}, Potsdam~\cite{Rottensteiner2014}, Vaihingen~\cite{Rottensteiner2014}, LoveDA~\cite{Wang2021}, iSAID~\cite{Zamir2019iSAID, Xia2018DOTA} and UAVid~\cite{LYU2020108}. We adopt a standard and widely used pixel-level evaluation metric, namely Mean Intersection over Union (mIoU) to quantitatively evaluate the segmentation performance. This metric first computes the Intersection over Union (IoU) for each class, defined as the ratio between the intersection and the union of the predicted region and the ground-truth region, and then averages the IoU values over all classes.

For a task with $N_C$ classes, mIoU is calculated as:
\begin{equation}
\mathrm{mIoU} = \frac{1}{N_C}\sum_{i=1}^{N_C}\frac{\mathrm{TP}_i}{\mathrm{TP}_i+\mathrm{FP}_i+\mathrm{FN}_i},
\label{eq:miou}
\end{equation}
where $\mathrm{TP}_i$, $\mathrm{FP}_i$, and $\mathrm{FN}_i$ denote the numbers of true positive, false positive, and false negative pixels for class $i$, respectively.

\subsection{Implementation Details}
\label{SEC_IMPLEMENTATION_DETAILS}
SDCI is implemented based on PyTorch. All experiments are conducted on a computer equipped with four NVIDIA GeForce RTX 3090 GPU devices. To ensure fair comparisons and to validate the generalization ability of our method, all experiments follow a fixed set of hyperparameters as described below.

The dual-branch architecture of SDCI consists of a semantic branch and a structural branch. For the semantic branch, we adopt the CLIP ViT-L/14 model released by OpenAI as the backbone. For the structural branch, we evaluate two pretrained DINO models, including DINO-v1 and DINO-v2. It is worth noting that, to ensure that the output feature dimensions can be efficiently aligned with other components in our framework and to evaluate the performance evolution across DINO generations on a fair benchmark, we consistently use a ViT-Base backbone for the structural branch. In all experiments, the parameters of these backbone networks are frozen.

In SDCI, CAF is implemented via direct additive injection (i.e., $\lambda_1=1$ in Eqs.~\ref{eq:caf_fuse_1} and \ref{eq:caf_fuse_2}). For Eqs.~\ref{eq:bcdg_sem_2}-\ref{eq:bcdg_str_1} in BCDR, the default settings are $K=30$ and $\tau=7$. For the diffusion process (i.e., Eqs.~\ref{eq:bcdg_br_1} and \ref{eq:bcdg_br_2}), the total number $T$ of diffusion steps is set to $40$, and the smoothing factor $\alpha$ is set to $0.9$. For Eqs.~\ref{eq:cscp_1}-\ref{eq:cscp_3} in CSCP, the balancing coefficients $\lambda_C$ and $\lambda_D$ are set to $1.0$ and $0.2$, the total variation regularization strength $\beta$ is set to $0.10$, and the adaptive edge-weight parameters $w_{\text{in}}$ and $w_{\text{cross}}$ are set to $1.0$ and $0.10$, respectively. All input images are uniformly cropped to $512 \times 512$ pixels. We adopt full-image inference, where the entire image is directly fed into the encoder.

\subsection{Prompt Settings}
\label{SEC_PROMPT}
To conduct a more rigorous and realistic evaluation of model performance, we design two prompt settings: \emph{original-label} and \emph{generalized-label} settings. 

\textbf{Original-Label (Ori) Setting:} We directly use the category names officially provided by each dataset as text prompts. This setting is mainly adopted to measure the benchmark performance under professional land-cover terminology.

\textbf{Generalized-Label (Gen) Setting:} This setting aims to systematically bridge the semantic gap between the official category names and the vocabulary space of CLIP. Official category names are often sub-optimal text prompts because they can be overly specialized (e.g., \emph{impervious surfaces}) or conceptually abstract (e.g., \emph{agriculture}), making them difficult to align with CLIP's natural-language-based understanding paradigm.

To produce these prompt settings, we design a two-stage prompt selection pipeline, as shown in Table~\ref{tab:gen_label}. First, for each original name, we construct a generalized label candidate set (Set of Gen). This Set of Gen is centered around the core visual concept and includes potential prompts spanning multiple semantic dimensions (e.g., constituent elements, functionality, and hierarchical relations). Subsequently, the final generalized label (Gen) is determined from Set of Gen. For example, the \emph{built-up} category in the GID dataset is visually almost entirely represented by well-defined individual buildings, and thus we choose \emph{buildings} to achieve the most direct and unambiguous visual correspondence. In contrast, the \emph{impervious surfaces} category in the Potsdam dataset constitutes a \emph{heterogeneous composite} composed of roads, parking lots, and plazas. Therefore, we select a semantically richer prompt set $\{\textit{ground}, \textit{road}, \textit{streets}\}$ to holistically capture its diverse visual semantics. By comparison, for the iSAID and UAVid datasets, since their category names (e.g., \emph{plane}, \emph{ship}) essentially correspond to concrete individual objects and are naturally aligned with the CLIP vocabulary space, we directly retain the official category names to avoid unnecessary semantic re-construction.

\begin{table*}[t]
\centering
\small
\setlength{\tabcolsep}{4pt}
\caption{Mapping details from the original (Ori) labels to the generalized (Gen) labels across different datasets. The Gen labels are manually designed to be more descriptive by decomposing complex or technical terms into concrete visual components, and are better aligned with the natural-language understanding of VLMs, thereby forming a more realistic evaluation benchmark.}
\label{tab:gen_label}
\begin{tabularx}{\textwidth}{@{} l l X X @{}}
\toprule
\textbf{Dataset} & \textbf{Ori} & \textbf{Candidate Set of Gen} & \textbf{Gen} \\
\midrule

\multirow{5}{*}{GID}
& built-up  & residential area, structures, buildings, architecture
            & buildings \\
& farmland  & cropland, agricultural land, field, cultivated land
            & agricultural land \\
& forest    & woodland, tree cover, mountain forest, canopy
            & mountain forest \\
& meadow    & grassland, greenspace, shrubs, flatland forest, low vegetation
            & flatland forest \\
& water     & river, lake, water body
            & water body \\
\midrule

\multirow{5}{*}{Potsdam}
& impervious surfaces & road, streets, ground, parking lot, sidewalk
                     & ground, road, streets, parking lot \\
& building            & buildings, man-made buildings, architecture, house
                     & buildings, man-made buildings \\
& low vegetation      & low vegetation, low-growing grassland, grass patches, lawn, grass vegetation
                     & low vegetation, low-growing grassland, grass patches, lawn, grass vegetation \\
& tree                & woods, trees, plants
                     & trees \\
& car                 & automobile, vehicles, transportation
                     & vehicles \\
\midrule

\multirow{5}{*}{Vaihingen}
& impervious surfaces & road, streets, ground, parking lot, sidewalk
                     & ground, road, streets, parking lot \\
& building            & building, man-made buildings, architecture, house
                     & building \\
& low vegetation      & low vegetation, low-growing grassland, grass
                     & low vegetation, low-growing grassland \\
& tree                & woods, tree, plants
                     & tree \\
& car                 & automobile, vehicle, transportation
                     & vehicle \\
\midrule

\multirow{6}{*}{LoveDA}
& building     & building, man-made buildings, architecture, house
              & building \\
& road         & road, streets
              & road \\
& water        & river, lake, water
              & water \\
& barren       & bare land, dry soil, sandy land
              & dry soil \\
& forest       & woodland, tree cover, forest
              & forest \\
& agriculture  & cropland, agricultural land, field, farmland
              & farmland \\
\bottomrule
\end{tabularx}
\end{table*}

\begin{table*}[t]
\centering
\small
\setlength{\tabcolsep}{3.6pt}
\renewcommand{\arraystretch}{1.15}
\caption{Quantitative comparison with existing methods on GID, Potsdam, Vaihingen, LoveDA, iSAID and UAVid. The accuracy values are mIoU scores (\%). Ori and Gen denote the prompt settings with Original Labels and Generalized Labels, respectively. GFLOPs are measured under a fixed input resolution of $512 \times 512$. SDCI-v1 and SDCI-v2 correspond to integrating DINO-v1 (ViT-B/8) and DINO-v2 (ViT-B/14) as the structural branch, respectively. Best results are in \textbf{bold}; second-best results among existing methods are \underline{underlined}. Blue numbers indicate the absolute improvement over the second-best method.}
\label{tab:main_results}
\begin{tabularx}{\textwidth}{@{} l *{11}{>{\centering\arraybackslash}X} >{\centering\arraybackslash}X @{}}
\toprule
\multirow{2}{*}{\textbf{Method}} 
& \multicolumn{2}{c}{\textbf{GID}} 
& \multicolumn{2}{c}{\textbf{Potsdam}} 
& \multicolumn{2}{c}{\textbf{Vaihingen}} 
& \multicolumn{2}{c}{\textbf{LoveDA}} 
& \multicolumn{1}{c}{\textbf{iSAID}} 
& \multicolumn{1}{c}{\textbf{UAVid}}
& \multirow{2}{*}{\textbf{Avg}} 
& \multirow{2}{*}{\textbf{GFLOPs}} \\
\cmidrule(lr){2-3}\cmidrule(lr){4-5}\cmidrule(lr){6-7}\cmidrule(lr){8-9}\cmidrule(lr){10-10}\cmidrule(lr){11-11}
& \textbf{Ori} & \textbf{Gen} & \textbf{Ori} & \textbf{Gen} & \textbf{Ori} & \textbf{Gen} & \textbf{Ori} & \textbf{Gen} & \textbf{Ori} & \textbf{Ori} &  &  \\
\midrule
MaskCLIP~\cite{Zhou2022MaskCLIP}   & 45.04 & 52.06 & 26.54 & 46.76 & 24.80 & 35.15 & 38.37 & 36.90 & 14.50 & 28.60 & 34.87 & 2944.88 \\
ClearCLIP~\cite{Lan2025ClearCLIP}  & 53.03 & 58.71 & 26.29 & 46.88 & 22.01 & 26.69 & 38.39 & 39.51 & 18.20 & 36.20 & 36.59 & 2953.55 \\
SCLIP~\cite{Wang2025SCLIP}       & 52.80 & 58.50 & 27.13 & 49.93 & 22.72 & 30.23 & 37.05 & 37.85 & 16.10 & 31.40 &36.37 & 2926.29 \\
CorrCLIP~\cite{Zhang2025CorrCLIP}   & 50.00 & 45.73 & 25.53 & 43.28 & 23.00 & 43.42 & 34.08 & 33.62 & 25.50 & 40.93 & 36.51 & 6424.47 \\
CLIPer~\cite{Sun2025CLIPer}     & 52.11 & \underline{66.85} & 25.96 & 45.22 & \underline{26.21} & \underline{43.97} & \underline{43.58} & \underline{43.84} & 27.36 & 40.90 & 41.60 & 4615.63 \\
NACLIP~\cite{Hajimiri2024NACLIP}     & 54.96 & 63.29 & \underline{33.47} & 32.03 & 20.93 & 33.14 & 32.29 & 32.20 & 31.57 & 37.91 & 37.18 & 242.96 \\
ResCLIP~\cite{Yang2024ResCLIP}    & 54.23 & 60.67 & 27.02 & \underline{53.31} & 23.39 & 38.32 & 39.88 & 40.85 & \underline{43.79} & 37.10 & \underline{41.86} & 1029.66 \\
CASS~\cite{Kim2025CASS}       & \underline{57.73} & 57.73 & 32.68 & 34.28 & 24.21 & 41.14 & 42.42 & 42.01 & 40.41 & 37.08 & 40.97 & 26943.79 \\
SegEarth-OV~\cite{Li2025SegEarth}   & 52.49 & 59.49 & 30.63 & 48.80 & 23.03 & 39.34 & 40.97 & 42.63 & 21.70 & \underline{42.50} & 40.16 & 1124.28 \\
\midrule
SDCI-v1 (Ours) 
& 58.88 \textcolor{blue}{(+1.15)} 
& \textbf{71.27} \textcolor{blue}{(+4.42)} 
& 32.42 \textcolor{blue}{(——)} 
& 58.16 \textcolor{blue}{(+4.85)} 
& 33.40 \textcolor{blue}{(+7.19)} 
& 46.21 \textcolor{blue}{(+2.24)} 
& \textbf{45.00} \textcolor{blue}{(+1.42)} 
& \textbf{45.30} \textcolor{blue}{(+1.46)} 
& 42.68 \textcolor{blue}{(——)} 
& \textbf{42.63} \textcolor{blue}{(+0.13)} 
& 47.60 \textcolor{blue}{(+5.74)} 
& 2022.72 \\
SDCI-v2 (Ours) 
& \textbf{59.58} \textcolor{blue}{(+1.85)} 
& 64.63 \textcolor{blue}{(——)} 
& \textbf{36.49} \textcolor{blue}{(+3.02)} 
& \textbf{58.33} \textcolor{blue}{(+5.02)} 
& \textbf{33.66} \textcolor{blue}{(+7.45)} 
& \textbf{50.68} \textcolor{blue}{(+6.71)} 
& 43.70 \textcolor{blue}{(+0.12)} 
& 42.65 \textcolor{blue}{(——)} 
& \textbf{46.43} \textcolor{blue}{(+2.64)} 
& 40.56 \textcolor{blue}{(——)} 
& \textbf{47.66} \textcolor{blue}{(+5.80)} 
& 1683.29 \\
\bottomrule
\end{tabularx}
\end{table*}

\subsection{Comparison with Other Methods}
As shown in Table~\ref{tab:main_results}, under the original-label setting, SDCI-v2 demonstrates outstanding overall performance, achieving an average mIoU of 47.66\% and outperforming all competing methods. For instance, on the GID dataset, SDCI-v2 (Ori) attains an mIoU of 59.58\%, surpassing the previous state-of-the-art method CASS (57.73\%). On the Vaihingen dataset, SDCI-v2 (33.66\%) significantly exceeds the second-best method CLIPer (26.21\%). Under the Gen setting, SDCI consistently maintains its advantage. For example, on the Potsdam dataset, the score of SDCI-v2 surges from 36.49\% in the Ori setting to 58.33\% in the Gen setting, which not only substantially surpasses its own benchmark performance but also exceeds ResCLIP (53.31\%) under the same setting by 5.02\%. This result indicates that our framework is able to flexibly handle different forms of text prompts.

In addition to its superior segmentation accuracy, our model is also acceptable in computational efficiency. Under a unified input resolution of $512 \times 512$, we evaluate the computational complexity of different SDCI variants and existing methods. Specifically, SDCI-v1 adopts the $8 \times 8$ patch strategy of DINO-v1, which significantly increases the feature sequence length and thus incurs a higher computational cost (2022.72~GFLOPs). In contrast, the best-performing SDCI-v2 (with a $14\times 14$ patch strategy) successfully reduces the computation to 1683.29~GFLOPs while preserving sufficient spatial details. This cost is only about $1/16$ of that of the highly compute-intensive model CASS (26943.79~GFLOPs), and it also avoids the severe performance degradation observed in the lightweight model NACLIP (242.96~GFLOPs), which achieves only 20.93\% mIoU on Vaihingen. Although SDCI-v2 requires slightly more computation than SegEarth-OV (1124.28~GFLOPs), this increase is justified given the substantial performance gains it delivers.

A particularly notable finding is that the significant improvement brought by the Gen setting from the fact that their textual descriptions closely match the pretraining distribution preferences and the demand for semantic concreteness of VLMs. Since models such as CLIP are trained on massive amounts of natural language from the Internet, their feature space is more sensitive to and better aligned with common, high-frequency, and concrete words, while responding more weakly to low-frequency, abstract, and specialized terms. Generalized labels effectively transform complex expert-level geoscience categories into intuitive visual-attribute descriptions, substantially reducing semantic ambiguity and enabling the high-level semantic features extracted by CLIP to be activated and retrieved more precisely.

\begin{table}[t]
\centering
\small
\setlength{\tabcolsep}{4pt}
\renewcommand{\arraystretch}{1.15}
\caption{Ablation study of each module in SDCI on the GID dataset based on mIoU (\%) scores.}
\label{tab:ablation_results}
\begin{tabularx}{\linewidth}{@{} l *{3}{>{\centering\arraybackslash}X} @{}}
\toprule
\textbf{Method} & \textbf{Ori} & \textbf{Gen} & \textbf{Set of Gen} \\
\midrule
Baseline 
& 49.75 
& 56.82 
& 57.58 \\
Baseline + DINO-v1 (Late Fusion) 
& 45.76 
& 55.60 
& 53.68 \\
Baseline + CAF 
& 53.40 (+3.65) 
& 61.86 (+5.04) 
& 58.22 (+0.64) \\
Row~3 + BCDR 
& 58.09 (+8.34) 
& 62.46 (+5.64) 
& 58.65 (+1.07) \\
Row~4 + CSCP (Ours, DINO-v1) 
& 58.88 (+9.13) 
& \textbf{71.27} (+14.45) 
& \textbf{64.17} (+6.59) \\
Row~4 + CSCP (Ours, DINO-v2) 
& \textbf{59.58} (+9.83) 
& 64.63 (+7.81) 
& 64.10 (+6.52) \\
\bottomrule
\end{tabularx}
\end{table}

\subsection{Ablation Study}
To ensure fair and meaningful comparisons, we adopt a strong CLIP-based baseline built upon multi-level feature fusion. This baseline follows the core ideas of CLIPer~\cite{Sun2025CLIPer} and clearCLIP~\cite{Lan2025ClearCLIP}, aiming to improve the spatial localization capability of standard CLIP by exploiting early-layer features. In addition, it simplifies the network structure of the final layer to reduce noise and better align visual features with text embeddings, thereby establishing a strong baseline that outperforms vanilla CLIP. Starting from this baseline, we conduct a progressive and incremental ablation study to quantify the contribution of each core component in SDCI. The results are reported in Table~\ref{tab:ablation_results}.

\begin{figure}[ht]
\begin{center}
\centerline{\includegraphics[width=3.3in]{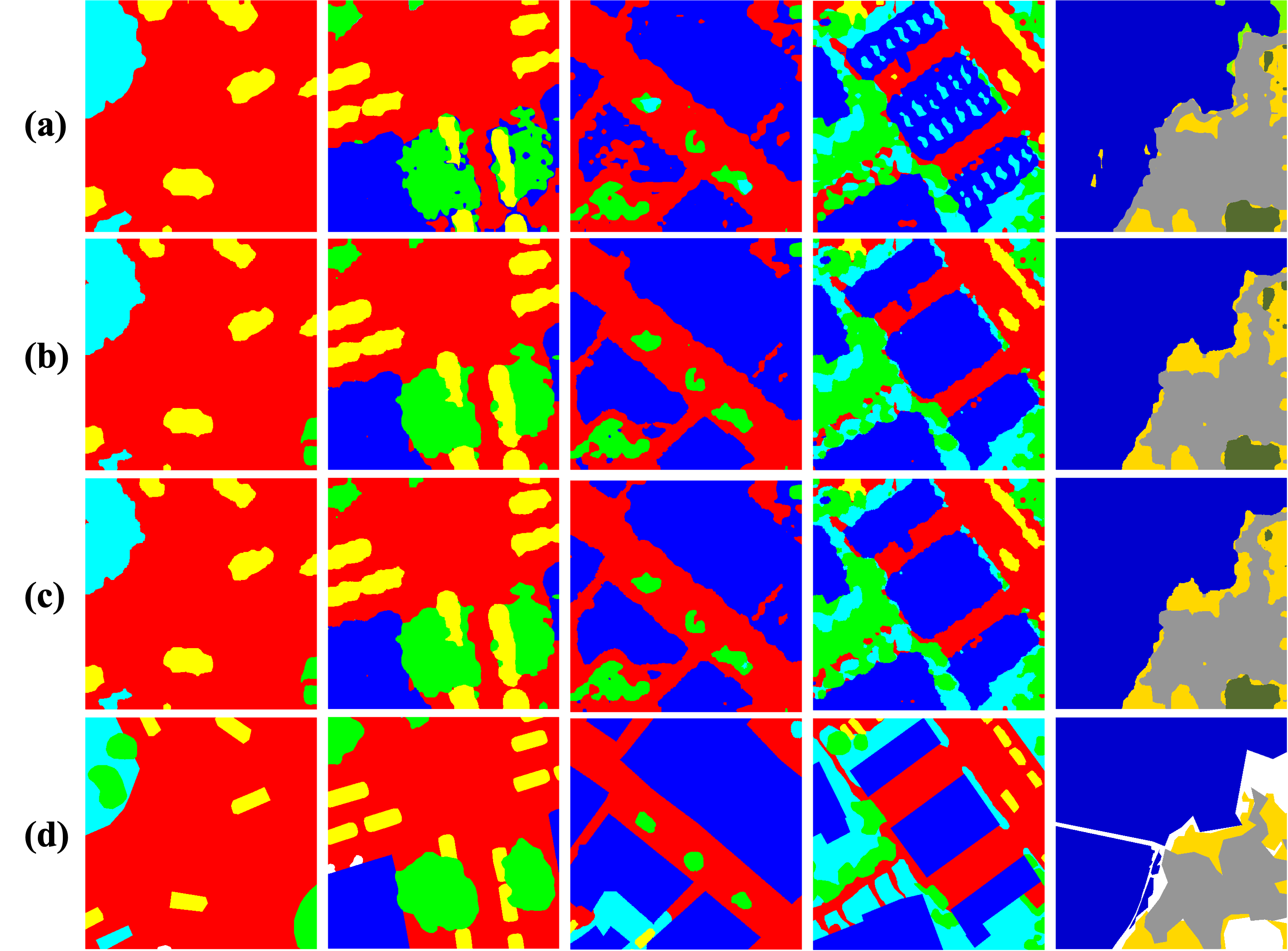}}
\caption{The impact of CSCP, (a) results of baseline + CAF + BCDR, (b) full model with CSCP (SDCI-v1), (c) full model with CSCP (SDCI-v2), (d) ground truth.}
\label{FIG_ABLATION}
\end{center}
\end{figure}

\paragraph{Effectiveness of Core Components.}
Under the original-label setting, our method exhibits a clear performance ladder. Starting from the baseline (49.75\%), the mIoU steadily increases to 59.58\% after progressively adding the CAF, BCDR, and CSCP modules. This consistent upward trend strongly demonstrates that each module is effective and makes a significant contribution. In the Gen and Set of Gen settings, when CSCP is integrated, the accuracy improves substantially, rising from 62.46\% (58.65\%) to 71.27\% (64.17\%). This highlights the significant contribution of CSCP and demonstrates that the geometric constraints induced by the superpixel structure are highly effective. As shown in figure~\ref{FIG_ABLATION}, CSCP helps preserve stable segmentation boundaries by leveraging intrinsic geometric cues.

\paragraph{Necessity of Simple Late Fusion Strategy.}
An insightful phenomenon appears in the second row in Table~\ref{tab:ablation_results}. When adopting a simple late fusion strategy---i.e., directly summing the semantic scores of DINO and CLIP with fixed weights---the performance under the Ori setting drops from 49.75\% to 45.76\%. We attribute this to the fact that although DINO's raw attention maps contain rich structural cues, they also include substantial structural noise irrelevant to the target semantics. Without the semantic guidance provided by CAF and BCDR, directly merging DINO with CLIP can interfere with accurate semantic discrimination. This performance degradation further highlights the necessity of CAF and BCDR for early and deep cross-model interaction.

\paragraph{Analysis of Structural Branch Selection.}
Within our full framework, comparing different DINO variants reveals the importance of spatial consistency. As shown in Table~\ref{tab:main_results}, both SDCI-v1 and SDCI-v2 obtain higher accuracy than existing methods, the SDCI-v2 which is equipped with DINO-v2 gets slightly higher accuracy than SDCI-v1. Via the analysis on the detailed architectures of DINO and CLIP, we give the following two reasons. First, it is crucial to strike the best trade-off between semantic discriminability and spatial granularity: while DINO-v1 ($8\times 8$) preserves abundant fine-grained texture details, DINO-v2 ($14\times 14$) benefits from stronger pretrained representations and excels at object-level visual modeling. The notable gain in feature discriminability outweighs the marginal impact caused by reduced spatial resolution. Second, since the baseline CLIP model also adopts a $14\times 14$ patch size, DINO-v2 naturally aligns with CLIP feature maps on the spatial grid. This point-to-point correspondence completely avoids the up/down-sampling operations required by DINO-v1 during cross-modal fusion, thereby eliminating spatial misalignment introduced by interpolations. 

\paragraph{Improved Semantic Generalization.}
As shown in Table~\ref{tab:ablation_results}, under the generalized-label setting, our full method (71.27\%) achieves a more substantial improvement over the baseline (56.82\%), yielding a gain of +14.45\%. Moreover, as shown in Table~\ref{tab:main_results}, our SDCI-v1 obtains the highest accuracy than existing methods on all the datasets except for Potsdam and iSAID, and our SDCI-v2 obtains higher accuracy than existing methods on all the datasets except for UAVid in the original-label setting. This demonstrates the strong robustness of our framework with respect to different quality of text prompts. 

\paragraph{Gen vs. Candidate Set of Gen.}
By comparing the performance under the generalized-label (Gen) setting and the ``Set of Gen" settings, we observe a consistent trend: using our carefully selected best prompts (Gen) generally outperforms using the entire candidate set. For example, our final model (SDCI-v2) achieves an mIoU of 71.27\% under the Gen setting, higher than 64.17\% under the ``Set of Gen" setting. This finding provides an important insight that \emph{the quality of text prompts matters more than their quantity}. Feeding a candidate set with imprecise prompts as input can instead introduce semantic noise, dilute the core visual concept, and consequently interfere with the model predictions. Notably, when using the ``Set of Gen" setting, the text embedding of each text prompt in the set are derived. Then, all the embeddings are averaged to produce the finally text embedding of each category.

\section{Conclusion}
This paper introduces a novel training-free multi-stage collaborative inference framework, termed SDCI, which aims to deeply fuse the semantic knowledge of CLIP with the structural knowledge of DINO, while further exploiting low-level geometric cues to tackle the challenge of OVSS in remote sensing imagery. Extensive experiments on multiple challenging benchmarks demonstrate that SDCI consistently outperforms existing methods under a training-free setting. Ablation studies further validate the positive effects of each module in SDCI and provide detailed analyses on the choice of the structural branch, and the improvement in semantic generalization. Despite the superior accuracy achieved by SDCI, it is undeniable that current training-free OVSS methods still lag behind traditional fully supervised approaches trained on specific datasets. Future work will focus on narrowing this performance gap. 

\begin{acks}
To Robert, for the bagels and explaining CMYK and color spaces.
\end{acks}

\bibliographystyle{ACM-Reference-Format}
\bibliography{SDCI}

@article{Shi2000NCuts,
	author = {Jianbo Shi and Malik, J.},
	doi = {10.1109/34.868688},
	journal = {IEEE Transactions on Pattern Analysis and Machine Intelligence},
	number = {8},
	pages = {888-905},
	title = {Normalized cuts and image segmentation},
	volume = {22},
	year = {2000}}

@inproceedings{Zhong2025OmniSAM,
	author = {Zhong, Ding and Zheng, Xu and Liao, Chenfei and Lyu, Yuanhuiyi and Chen, Jialei and Wu, Shengyang and Zhang, Linfeng and Hu, Xuming},
	booktitle = {Proceedings of the {IEEE}/{CVF} {International} {Conference} on {Computer} {Vision} ({ICCV})},
	month = oct,
	pages = {23892--23901},
	title = {{OmniSAM}: {Omnidirectional} {Segment} {Anything} {Model} for {UDA} in {Panoramic} {Semantic} {Segmentation}},
	year = {2025}}

@inproceedings{Shi2025Trident,
	author = {Shi, Yuheng and Dong, Minjing and Xu, Chang},
	booktitle = {Proceedings of the {IEEE}/{CVF} {International} {Conference} on {Computer} {Vision} ({ICCV})},
	month = oct,
	pages = {23487--23497},
	title = {Harnessing {Vision} {Foundation} {Models} for {High}-{Performance}, {Training}-{Free} {Open} {Vocabulary} {Segmentation}},
	year = {2025}}

@inproceedings{Ge2025CRTNet,
	author = {Ge, Jiannan and Xie, Lingxi and Xie, Hongtao and Li, Pandeng and Liu, Sun-Ao and Zhang, Xiaopeng and Tian, Qi and Zhang, Yongdong},
	booktitle = {Proceedings of the {IEEE}/{CVF} {International} {Conference} on {Computer} {Vision} ({ICCV})},
	month = oct,
	pages = {24034--24044},
	title = {{CLIP}-{Adapted} {Region}-to-{Text} {Learning} for {Generative} {Open}-{Vocabulary} {Semantic} {Segmentation}},
	year = {2025}}

@article{Tong2020GID,
	author = {Xin-Yi Tong and Gui-Song Xia and Qikai Lu and Huanfeng Shen and Shengyang Li and Shucheng You and Liangpei Zhang},
	doi = {https://doi.org/10.1016/j.rse.2019.111322},
	issn = {0034-4257},
	journal = {Remote Sensing of Environment},
	pages = {111322},
	title = {Land-cover classification with high-resolution remote sensing images using transferable deep models},
	url = {https://www.sciencedirect.com/science/article/pii/S0034425719303414},
	volume = {237},
	year = {2020}}

@article{Chambolle2011PDHG,
	author = {Chambolle, Antonin and Pock, Thomas},
	date = {2011/05/01},
	doi = {10.1007/s10851-010-0251-1},
	id = {Chambolle2011},
	isbn = {1573-7683},
	journal = {Journal of Mathematical Imaging and Vision},
	number = {1},
	pages = {120--145},
	title = {A First-Order Primal-Dual Algorithm for Convex Problems with Applications to Imaging},
	url = {https://doi.org/10.1007/s10851-010-0251-1},
	volume = {40},
	year = {2011}}

@misc{Simeoni2025DINO-v3,
	author = {Sim{\'e}oni, Oriane and Vo, Huy V. and Seitzer, Maximilian and Baldassarre, Federico and Oquab, Maxime and Jose, Cijo and Khalidov, Vasil and Szafraniec, Marc and Yi, Seungeun and Ramamonjisoa, Micha{\"e}l and Massa, Francisco and Haziza, Daniel and Wehrstedt, Luca and Wang, Jianyuan and Darcet, Timoth{\'e}e and Moutakanni, Th{\'e}o and Sentana, Leonel and Roberts, Claire and Vedaldi, Andrea and Tolan, Jamie and Brandt, John and Couprie, Camille and Mairal, Julien and J{\'e}gou, Herv{\'e} and Labatut, Patrick and Bojanowski, Piotr},
	note = {\_eprint: 2508.10104},
	title = {{DINOv3}},
	url = {https://arxiv.org/abs/2508.10104},
	year = {2025}}

@article{Wang2023TokenCut,
	author = {Wang, Yangtao and Shen, Xi and Yuan, Yuan and Du, Yuming and Li, Maomao and Hu, Shell Xu and Crowley, James L. and Vaufreydaz, Dominique},
	doi = {10.1109/TPAMI.2023.3305122},
	issn = {1939-3539},
	journal = {IEEE Transactions on Pattern Analysis and Machine Intelligence},
	month = dec,
	number = {12},
	pages = {15790--15801},
	title = {{TokenCut}: {Segmenting} {Objects} in {Images} and {Videos} {With} {Self}-{Supervised} {Transformer} and {Normalized} {Cut}},
	volume = {45},
	year = {2023}}

@inproceedings{Lan2024ProxyCLIP,
	author = {Lan, Mengcheng and Chen, Chaofeng and Ke, Yiping and Wang, Xinjiang and Feng, Litong and Zhang, Wayne},
	booktitle = {European {Conference} on {Computer} {Vision}},
	pages = {70--88},
	publisher = {Springer},
	title = {Proxyclip: {Proxy} attention improves clip for open-vocabulary segmentation},
	year = {2024}}

@inproceedings{Sun2025CLIPer,
	author = {Sun, Lin and Cao, Jiale and Xie, Jin and Jiang, Xiaoheng and Pang, Yanwei},
	booktitle = {Proceedings of the {IEEE}/{CVF} {International} {Conference} on {Computer} {Vision} ({ICCV})},
	month = oct,
	pages = {23199--23209},
	title = {{CLIPer}: {Hierarchically} {Improving} {Spatial} {Representation} of {CLIP} for {Open}-{Vocabulary} {Semantic} {Segmentation}},
	year = {2025}}

@inproceedings{Abbasi2025ClipOscope,
	author = {Abbasi, Reza and Nazari, Ali and Sefid, Aminreza and Banayeeanzade, Mohammadali and Rohban, Mohammad Hossein and Baghshah, Mahdieh Soleymani},
	booktitle = {Proceedings of the {IEEE}/{CVF} {Conference} on {Computer} {Vision} and {Pattern} {Recognition} ({CVPR})},
	month = jun,
	pages = {9308--9317},
	title = {{CLIP} {Under} the {Microscope}: {A} {Fine}-{Grained} {Analysis} of {Multi}-{Object} {Representation}},
	year = {2025}}

@inproceedings{Barsellotti2024bFreeDA,
	author = {Barsellotti, Luca and Amoroso, Roberto and Cornia, Marcella and Baraldi, Lorenzo and Cucchiara, Rita},
	booktitle = {Proceedings of the {IEEE}/{CVF} {Conference} on {Computer} {Vision} and {Pattern} {Recognition} ({CVPR})},
	month = jun,
	pages = {3689--3698},
	title = {Training-{Free} {Open}-{Vocabulary} {Segmentation} with {Offline} {Diffusion}-{Augmented} {Prototype} {Generation}},
	year = {2024}}

@inproceedings{Barsellotti2025Talk2DINO,
	author = {Barsellotti, Luca and Bianchi, Lorenzo and Messina, Nicola and Carrara, Fabio and Cornia, Marcella and Baraldi, Lorenzo and Falchi, Fabrizio and Cucchiara, Rita},
	booktitle = {Proceedings of the {IEEE}/{CVF} {International} {Conference} on {Computer} {Vision} ({ICCV})},
	month = oct,
	pages = {22025--22035},
	title = {Talking to {DINO}: {Bridging} {Self}-{Supervised} {Vision} {Backbones} with {Language} for {Open}-{Vocabulary} {Segmentation}},
	year = {2025}}

@article{Hajimiri2024NACLIP,
	author = {Hajimiri, Sina and Ayed, Ismail Ben and Dolz, Jos{\'e}},
	journal = {2025 IEEE/CVF Winter Conference on Applications of Computer Vision (WACV)},
	pages = {5061--5071},
	title = {Pay {Attention} to {Your} {Neighbours}: {Training}-{Free} {Open}-{Vocabulary} {Semantic} {Segmentation}},
	url = {https://api.semanticscholar.org/CorpusID:269137404},
	year = {2025}}

@inproceedings{Kang2024LaVG,
	address = {Berlin, Heidelberg},
	author = {Kang, Dahyun and Cho, Minsu},
	booktitle = {Computer {Vision} -- {ECCV} 2024: 18th {European} {Conference}, {Milan}, {Italy}, {September} 29--{October} 4, 2024, {Proceedings}, {Part} {XLI}},
	doi = {10.1007/978-3-031-72940-9_9},
	isbn = {978-3-031-72939-3},
	pages = {143--164},
	publisher = {Springer-Verlag},
	title = {In {Defense} of {Lazy} {Visual} {Grounding} for {Open}-{Vocabulary} {Semantic} {Segmentation}},
	url = {https://doi.org/10.1007/978-3-031-72940-9_9},
	year = {2024}}

@inproceedings{Karazija2025OVDiff,
	address = {Cham},
	author = {Karazija, Laurynas and Laina, Iro and Vedaldi, Andrea and Rupprecht, Christian},
	booktitle = {Computer {Vision} -- {ECCV} 2024},
	editor = {Leonardis, Ale{\v s} and Ricci, Elisa and Roth, Stefan and Russakovsky, Olga and Sattler, Torsten and Varol, G{\"u}l},
	isbn = {978-3-031-72652-1},
	pages = {299--317},
	publisher = {Springer Nature Switzerland},
	title = {Diffusion {Models} for {Open}-{Vocabulary} {Segmentation}},
	year = {2025}}

@article{Kirillov2023SAM,
	author = {Kirillov, Alexander and Mintun, Eric and Ravi, Nikhila and Mao, Hanzi and Rolland, Chlo{\'e} and Gustafson, Laura and Xiao, Tete and Whitehead, Spencer and Berg, Alexander C. and Lo, Wan-Yen and Doll{\'a}r, Piotr and Girshick, Ross B.},
	journal = {2023 IEEE/CVF International Conference on Computer Vision (ICCV)},
	pages = {3992--4003},
	title = {Segment {Anything}},
	url = {https://api.semanticscholar.org/CorpusID:257952310},
	year = {2023}}

@inproceedings{Lan2025ClearCLIP,
	address = {Cham},
	author = {Lan, Mengcheng and Chen, Chaofeng and Ke, Yiping and Wang, Xinjiang and Feng, Litong and Zhang, Wayne},
	booktitle = {Computer {Vision} -- {ECCV} 2024},
	editor = {Leonardis, Ale{\v s} and Ricci, Elisa and Roth, Stefan and Russakovsky, Olga and Sattler, Torsten and Varol, G{\"u}l},
	isbn = {978-3-031-72970-6},
	pages = {143--160},
	publisher = {Springer Nature Switzerland},
	title = {{ClearCLIP}: {Decomposing} {CLIP} {Representations} for {Dense} {Vision}-{Language} {Inference}},
	year = {2025}}

@inproceedings{Shao2025CLIPtrase,
	address = {Cham},
	author = {Shao, Tong and Tian, Zhuotao and Zhao, Hang and Su, Jingyong},
	booktitle = {Computer {Vision} -- {ECCV} 2024},
	editor = {Leonardis, Ale{\v s} and Ricci, Elisa and Roth, Stefan and Russakovsky, Olga and Sattler, Torsten and Varol, G{\"u}l},
	isbn = {978-3-031-73016-0},
	pages = {139--156},
	publisher = {Springer Nature Switzerland},
	title = {Explore the {Potential} of {CLIP} for {Training}-{Free} {Open} {Vocabulary} {Semantic} {Segmentation}},
	year = {2025}}

@inproceedings{Shin2022ReCo,
	author = {Shin, Gyungin and Xie, Weidi and Albanie, Samuel},
	booktitle = {Advances in {Neural} {Information} {Processing} {Systems}},
	editor = {Koyejo, S. and Mohamed, S. and Agarwal, A. and Belgrave, D. and Cho, K. and Oh, A.},
	pages = {33754--33767},
	publisher = {Curran Associates, Inc.},
	title = {{ReCo}: {Retrieve} and {Co}-segment for {Zero}-shot {Transfer}},
	url = {https://proceedings.neurips.cc/paper_files/paper/2022/file/daabe43c3e1d06980aa23880bfbe1f45-Paper-Conference.pdf},
	volume = {35},
	year = {2022}}

@article{Simeoni2022,
	author = {Sim{\'e}oni, Oriane and Sekkat, Chlo'e and Puy, Gilles and Vobeck{\'y}, Anton{\'\i}n and Zablocki, {\'E}loi and P'erez, Patrick},
	journal = {2023 IEEE/CVF Conference on Computer Vision and Pattern Recognition (CVPR)},
	pages = {3176--3186},
	title = {Unsupervised {Object} {Localization}: {Observing} the {Background} to {Discover} {Objects}},
	url = {https://api.semanticscholar.org/CorpusID:254685799},
	year = {2022}}

@article{Sun2023CaR,
	author = {Sun, Shuyang and Li, Runjia and Torr, Philip and Gu, Xiuye and Li, Siyang},
	journal = {2024 IEEE/CVF Conference on Computer Vision and Pattern Recognition (CVPR)},
	pages = {13171--13182},
	title = {{CLIP} as {RNN}: {Segment} {Countless} {Visual} {Concepts} without {Training} {Endeavor}},
	url = {https://api.semanticscholar.org/CorpusID:266191302},
	year = {2023}}

@inproceedings{Wang2025SCLIP,
	address = {Cham},
	author = {Wang, Feng and Mei, Jieru and Yuille, Alan},
	booktitle = {Computer {Vision} -- {ECCV} 2024},
	editor = {Leonardis, Ale{\v s} and Ricci, Elisa and Roth, Stefan and Russakovsky, Olga and Sattler, Torsten and Varol, G{\"u}l},
	isbn = {978-3-031-72664-4},
	pages = {315--332},
	publisher = {Springer Nature Switzerland},
	title = {{SCLIP}: {Rethinking} {Self}-{Attention} for {Dense} {Vision}-{Language} {Inference}},
	year = {2025}}

@article{Wang2022,
	author = {Wang, Yangtao and Shen, X. I. and Hu, Shell Xu and Yuan, Yuan and Crowley, James L. and Vaufreydaz, Dominique},
	journal = {2022 IEEE/CVF Conference on Computer Vision and Pattern Recognition (CVPR)},
	pages = {14523--14533},
	title = {Self-{Supervised} {Transformers} for {Unsupervised} {Object} {Discovery} using {Normalized} {Cut}},
	url = {https://api.semanticscholar.org/CorpusID:247058696},
	year = {2022}}

@inproceedings{Wysoczanska2025CLIPDINOiser,
	address = {Cham},
	author = {Wysocza{\'n}ska, Monika and Sim{\'e}oni, Oriane and Ramamonjisoa, Micha{\"e}l and Bursuc, Andrei and Trzci{\'n}ski, Tomasz and P{\'e}rez, Patrick},
	booktitle = {Computer {Vision} -- {ECCV} 2024},
	editor = {Leonardis, Ale{\v s} and Ricci, Elisa and Roth, Stefan and Russakovsky, Olga and Sattler, Torsten and Varol, G{\"u}l},
	isbn = {978-3-031-73030-6},
	pages = {320--337},
	publisher = {Springer Nature Switzerland},
	title = {{CLIP}-{DINOiser}: {Teaching} {CLIP} a {Few} {DINO} {Tricks} for {Open}-{Vocabulary} {Semantic} {Segmentation}},
	year = {2025}}

@article{Yang2024ResCLIP,
	author = {Yang, Yuhang and Deng, Jinhong and Li, Wen and Duan, Lixin},
	journal = {2025 IEEE/CVF Conference on Computer Vision and Pattern Recognition (CVPR)},
	pages = {29968--29978},
	title = {{ResCLIP}: {Residual} {Attention} for {Training}-free {Dense} {Vision}-language {Inference}},
	url = {https://api.semanticscholar.org/CorpusID:274234839},
	year = {2025}}

@inproceedings{Zhang2025CorrCLIP,
	author = {Zhang, Dengke and Liu, Fagui and Tang, Quan},
	booktitle = {Proceedings of the {IEEE}/{CVF} {International} {Conference} on {Computer} {Vision} ({ICCV})},
	month = oct,
	pages = {24677--24687},
	title = {{CorrCLIP}: {Reconstructing} {Patch} {Correlations} in {CLIP} for {Open}-{Vocabulary} {Semantic} {Segmentation}},
	year = {2025}}

@inproceedings{Zhou2022MaskCLIP,
	address = {Berlin, Heidelberg},
	author = {Zhou, Chong and Loy, Chen Change and Dai, Bo},
	booktitle = {Computer {Vision} -- {ECCV} 2022: 17th {European} {Conference}, {Tel} {Aviv}, {Israel}, {October} 23--27, 2022, {Proceedings}, {Part} {XXVIII}},
	doi = {10.1007/978-3-031-19815-1_40},
	isbn = {978-3-031-19814-4},
	pages = {696--712},
	publisher = {Springer-Verlag},
	title = {Extract {Free} {Dense} {Labels} from {CLIP}},
	url = {https://doi.org/10.1007/978-3-031-19815-1_40},
	year = {2022}}

@article{Oquab2024DINO-v2,
	author = {Oquab, Maxime and Darcet, Timoth{\'e}e and Moutakanni, Th{\'e}o and Vo, Huy V. and Szafraniec, Marc and Khalidov, Vasil and Fernandez, Pierre and HAZIZA, Daniel and Massa, Francisco and El-Nouby, Alaaeldin and Assran, Mido and Ballas, Nicolas and Galuba, Wojciech and Howes, Russell and Huang, Po-Yao and Li, Shang-Wen and Misra, Ishan and Rabbat, Michael and Sharma, Vasu and Synnaeve, Gabriel and Xu, Hu and Jegou, Herve and Mairal, Julien and Labatut, Patrick and Joulin, Armand and Bojanowski, Piotr},
	issn = {2835-8856},
	journal = {Transactions on Machine Learning Research},
	title = {{DINOv2}: {Learning} {Robust} {Visual} {Features} without {Supervision}},
	url = {https://openreview.net/forum?id=a68SUt6zFt},
	year = {2024}}

@inproceedings{Simeoni2021,
	author = {Sim{\'e}oni, Oriane and Puy, Gilles and Vo, Huy V. and Roburin, Simon and Gidaris, Spyros and Bursuc, Andrei and P{\'e}rez, Patrick and Marlet, Renaud and Ponce, Jean},
	booktitle = {32nd {British} {Machine} {Vision} {Conference} 2021, {BMVC} 2021, {Online}, {November} 22-25, 2021},
	pages = {310},
	publisher = {BMVA Press},
	title = {Localizing {Objects} with {Self}-supervised {Transformers} and no {Labels}},
	url = {https://www.bmvc2021-virtualconference.com/assets/papers/1339.pdf},
	year = {2021}}

@inproceedings{Barsellotti2024aFOSSIL,
	author = {Barsellotti, Luca and Amoroso, Roberto and Baraldi, Lorenzo and Cucchiara, Rita},
	booktitle = {{IEEE}/{CVF} {Winter} {Conference} on {Applications} of {Computer} {Vision}, {WACV} 2024, {Waikoloa}, {HI}, {USA}, {January} 3-8, 2024},
	doi = {10.1109/WACV57701.2024.00149},
	pages = {1453--1462},
	publisher = {IEEE},
	title = {{FOSSIL}: {Free} {Open}-{Vocabulary} {Semantic} {Segmentation} through {Synthetic} {References} {Retrieval}},
	url = {https://doi.org/10.1109/WACV57701.2024.00149},
	year = {2024}}

@inproceedings{Kim2025CASS,
	author = {Kim, Chanyoung and Ju, Dayun and Han, Woojung and Yang, Ming-Hsuan and Hwang, Seong Jae},
	booktitle = {{IEEE}/{CVF} {Conference} on {Computer} {Vision} and {Pattern} {Recognition}, {CVPR} 2025, {Nashville}, {TN}, {USA}, {June} 11-15, 2025},
	doi = {10.1109/CVPR52734.2025.01400},
	pages = {15033--15042},
	publisher = {Computer Vision Foundation / IEEE},
	title = {Distilling {Spectral} {Graph} for {Object}-{Context} {Aware} {Open}-{Vocabulary} {Semantic} {Segmentation}},
	url = {https://openaccess.thecvf.com/content/CVPR2025/html/Kim\_Distilling\_Spectral\_Graph\_for\_Object-Context\_Aware\_Open-Vocabulary\_Semantic\_Segmentation\_CVPR\_2025\_paper.html},
	year = {2025}}

@inproceedings{Rombach2022LDM,
	author = {Rombach, Robin and Blattmann, Andreas and Lorenz, Dominik and Esser, Patrick and Ommer, Bj{\"o}rn},
	booktitle = {Proceedings of the {IEEE}/{CVF} {Conference} on {Computer} {Vision} and {Pattern} {Recognition} ({CVPR})},
	month = jun,
	pages = {10684--10695},
	title = {High-{Resolution} {Image} {Synthesis} {With} {Latent} {Diffusion} {Models}},
	year = {2022}}

@inproceedings{Wang2021,
	author = {Wang, Junjue and Zheng, Zhuo and Ma, Ailong and Lu, Xiaoyan and Zhong, Yanfei},
	booktitle = {Proceedings of the {Neural} {Information} {Processing} {Systems} {Track} on {Datasets} and {Benchmarks}},
	editor = {Vanschoren, J. and Yeung, S.},
	publisher = {Curran Associates, Inc.},
	title = {{LoveDA}: {A} {Remote} {Sensing} {Land}-{Cover} {Dataset} for {Domain} {Adaptive} {Semantic} {Segmentation}},
	url = {https://datasets-benchmarks-proceedings.neurips.cc/paper_files/paper/2021/file/4e732ced3463d06de0ca9a15b6153677-Paper-round2.pdf},
	volume = {1},
	year = {2021}}

@article{Rottensteiner2014,
	author = {Rottensteiner, Franz and Sohn, Gunho and Gerke, Markus and Wegner, Jan Dirk},
	doi = {10.1016/j.isprsjprs.2013.10.004},
	journal = {ISPRS Journal of Photogrammetry and Remote Sensing},
	pages = {143--171},
	publisher = {Elsevier},
	title = {The {ISPRS} benchmark on urban object detection and {3D} building reconstruction},
	volume = {93},
	year = {2014}}

@article{Felzenszwalb2004,
	author = {Felzenszwalb, Pedro F. and Huttenlocher, Daniel P.},
	doi = {10.1023/B:VISI.0000022288.19776.77},
	issn = {1573-1405},
	journal = {International Journal of Computer Vision},
	month = sep,
	number = {2},
	pages = {167--181},
	title = {Efficient {Graph}-{Based} {Image} {Segmentation}},
	url = {https://doi.org/10.1023/B:VISI.0000022288.19776.77},
	volume = {59},
	year = {2004}}

@inproceedings{Li2025SegEarth,
	author = {Li, Kaiyu and Liu, Ruixun and Cao, Xiangyong and Bai, Xueru and Zhou, Feng and Meng, Deyu and Wang, Zhi},
	booktitle = {Proceedings of the {Computer} {Vision} and {Pattern} {Recognition} {Conference} ({CVPR})},
	month = jun,
	pages = {10545--10556},
	title = {{SegEarth}-{OV}: {Towards} {Training}-{Free} {Open}-{Vocabulary} {Segmentation} for {Remote} {Sensing} {Images}},
	year = {2025}}

@inproceedings{Fu2024SimFeatUp,
	author = {Fu, Stephanie and Hamilton, Mark and Brandt, Laura E. and Feldmann, Axel and Zhang, Zhoutong and Freeman, William T.},
	booktitle = {The {Twelfth} {International} {Conference} on {Learning} {Representations}},
	title = {{FeatUp}: {A} {Model}-{Agnostic} {Framework} for {Features} at {Any} {Resolution}},
	url = {https://openreview.net/forum?id=GkJiNn2QDF},
	year = {2024}}

@inproceedings{Radford2021CLIP,
	author = {Radford, Alec and Kim, Jong Wook and Hallacy, Chris and Ramesh, Aditya and Goh, Gabriel and Agarwal, Sandhini and Sastry, Girish and Askell, Amanda and Mishkin, Pamela and Clark, Jack and Krueger, Gretchen and Sutskever, Ilya},
	booktitle = {Proceedings of the 38th International Conference on Machine Learning},
	editor = {Meila, Marina and Zhang, Tong},
	month = {18--24 Jul},
	pages = {8748--8763},
	publisher = {PMLR},
	series = {Proceedings of Machine Learning Research},
	title = {Learning Transferable Visual Models From Natural Language Supervision},
	url = {https://proceedings.mlr.press/v139/radford21a.html},
	volume = {139},
	year = {2021}}

@inproceedings{Caron2021DINO,
	author = {Caron, Mathilde and Touvron, Hugo and Misra, Ishan and Jegou, Herv{\'e} and Mairal, Julien and Bojanowski, Piotr and Joulin, Armand},
	booktitle = {2021 IEEE/CVF International Conference on Computer Vision (ICCV)},
	doi = {10.1109/ICCV48922.2021.00951},
	pages = {9630-9640},
	title = {Emerging Properties in Self-Supervised Vision Transformers},
	year = {2021}}

@inproceedings{Wang2023CutLER,
	author = {Wang, Xudong and Girdhar, Rohit and Yu, Stella X and Misra, Ishan},
	booktitle = {Proceedings of the IEEE/CVF Conference on Computer Vision and Pattern Recognition},
	pages = {3124--3134},
	title = {Cut and learn for unsupervised object detection and instance segmentation},
	year = {2023}}

@inproceedings{Wang2024VCutLER,
	address = {Los Alamitos, CA, USA},
	author = {Wang, Xudong and Misra, Ishan and Zeng, Ziyun and Girdhar, Rohit and Darrell, Trevor},
	booktitle = {2024 IEEE/CVF Conference on Computer Vision and Pattern Recognition (CVPR)},
	doi = {10.1109/CVPR52733.2024.02147},
	month = Jun,
	pages = {22755-22764},
	publisher = {IEEE Computer Society},
	title = {{ VideoCutLER: Surprisingly Simple Unsupervised Video Instance Segmentation }},
	url = {https://doi.ieeecomputersociety.org/10.1109/CVPR52733.2024.02147},
	year = {2024}}

@article{Ravi2024sam2,
	author = {Ravi, Nikhila and Gabeur, Valentin and Hu, Yuan-Ting and Hu, Ronghang and Ryali, Chaitanya and Ma, Tengyu and Khedr, Haitham and R{\"a}dle, Roman and Rolland, Chloe and Gustafson, Laura and Mintun, Eric and Pan, Junting and Alwala, Kalyan Vasudev and Carion, Nicolas and Wu, Chao-Yuan and Girshick, Ross and Doll{\'a}r, Piotr and Feichtenhofer, Christoph},
	journal = {arXiv preprint arXiv:2408.00714},
	title = {SAM 2: Segment Anything in Images and Videos},
	url = {https://arxiv.org/abs/2408.00714},
	year = {2024}}

@inproceedings{Zamir2019iSAID,
	author = {Waqas Zamir, Syed and Arora, Aditya and Gupta, Akshita and Khan, Salman and Sun, Guolei and Shahbaz Khan, Fahad and Zhu, Fan and Shao, Ling and Xia, Gui-Song and Bai, Xiang},
	booktitle = {Proceedings of the {IEEE} {Conference} on {Computer} {Vision} and {Pattern} {Recognition} {Workshops}},
	date = {2019},
	pages = {28--37},
	title = {{iSAID}: {A} {Large}-scale {Dataset} for {Instance} {Segmentation} in {Aerial} {Images}},
	year = {2019}}

@inproceedings{Xia2018DOTA,
	author = {Xia, Gui-Song and Bai, Xiang and Ding, Jian and Zhu, Zhen and Belongie, Serge and Luo, Jiebo and Datcu, Mihai and Pelillo, Marcello and Zhang, Liangpei},
	booktitle = {The {IEEE} {Conference} on {Computer} {Vision} and {Pattern} {Recognition} ({CVPR})},
	date = {2018-06},
	title = {{DOTA}: {A} {Large}-{Scale} {Dataset} for {Object} {Detection} in {Aerial} {Images}},
	year = {2018}}

@inproceedings{Zhang2025E-SAM,
	author = {Zhang, Weiming and Xiao, Dingwen and Chen, Lei and Wang, Lin},
	booktitle = {Proceedings of the IEEE/CVF International Conference on Computer Vision (ICCV)},
	month = {October},
	pages = {24688-24697},
	title = {E-SAM: Training-Free Segment Every Entity Model},
	year = {2025}}

@article{LYU2020108,
	author = {Ye Lyu and George Vosselman and Gui-Song Xia and Alper Yilmaz and Michael Ying Yang},
	doi = {https://doi.org/10.1016/j.isprsjprs.2020.05.009},
	issn = {0924-2716},
	journal = {ISPRS Journal of Photogrammetry and Remote Sensing},
	pages = {108-119},
	title = {UAVid: A semantic segmentation dataset for UAV imagery},
	url = {https://www.sciencedirect.com/science/article/pii/S0924271620301295},
	volume = {165},
	year = {2020}}

@inproceedings{Wang2025,
	author = {Wang, Zhengyang and Feng, Tingliang and Lyu, Fan and Shang, Fanhua and Feng, Wei and Wan, Liang},
	booktitle = {2025 IEEE/CVF Conference on Computer Vision and Pattern Recognition (CVPR)},
	doi = {10.1109/CVPR52734.2025.01882},
	pages = {20212-20222},
	title = {Dual Semantic Guidance for Open Vocabulary Semantic Segmentation},
	year = {2025}}

@inproceedings{Howlader_2026_WACV,
	author = {Howlader, Prantik and Nguyen-Canh, Hoang and Das, Srijan and Xu, Jingyi and Le, Hieu and Samaras, Dimitris},
	booktitle = {Proceedings of the IEEE/CVF Winter Conference on Applications of Computer Vision (WACV)},
	month = {March},
	pages = {5934-5944},
	title = {CORA: Consistency-Guided Semi-Supervised Framework for Reasoning Segmentation},
	year = {2026}}

@inproceedings{Wang2024PICLIP,
	author = {Wang, Jin and Zhang, Bingfeng and Pang, Jian and Chen, Honglong and Liu, Weifeng},
	booktitle = {2024 IEEE/CVF Conference on Computer Vision and Pattern Recognition (CVPR)},
	doi = {10.1109/CVPR52733.2024.00378},
	pages = {3941-3951},
	title = {Rethinking Prior Information Generation with CLIP for Few-Shot Segmentation},
	year = {2024}}

@inproceedings{Yoon_2026_WACV,
	author = {Yoon, Euihyun and Park, Taejin and Lee, Jaekoo},
	booktitle = {Proceedings of the IEEE/CVF Winter Conference on Applications of Computer Vision (WACV)},
	month = {March},
	pages = {6517-6526},
	title = {Training-Free Few-Shot Segmentation via Vision-Language Guided Prompting},
	year = {2026}}

@article{Rao2021DenseCLIPLD,
	author = {Yongming Rao and Wenliang Zhao and Guangyi Chen and Yansong Tang and Zheng Zhu and Guan Huang and Jie Zhou and Jiwen Lu},
	journal = {2022 IEEE/CVF Conference on Computer Vision and Pattern Recognition (CVPR)},
	pages = {18061-18070},
	title = {DenseCLIP: Language-Guided Dense Prediction with Context-Aware Prompting},
	url = {https://api.semanticscholar.org/CorpusID:244800733},
	year = {2021}}

@inproceedings{hamilton2022unsupervised,
	author = {Mark Hamilton and Zhoutong Zhang and Bharath Hariharan and Noah Snavely and William T. Freeman},
	booktitle = {International Conference on Learning Representations},
	title = {Unsupervised Semantic Segmentation by Distilling Feature Correspondences},
	url = {https://openreview.net/forum?id=SaKO6z6Hl0c},
	year = {2022}}

@article{zhang2023personalize,
	author = {Zhang, Renrui and Jiang, Zhengkai and Guo, Ziyu and Yan, Shilin and Pan, Junting and Dong, Hao and Gao, Peng and Li, Hongsheng},
	booktitle = {International Conference on Learning Representations},
	title = {Personalize Segment Anything Model with One Shot},
	year = {2024}}

\end{document}